\documentclass{article}

\usepackage{PRIMEarxiv}

\usepackage[utf8]{inputenc} 
\usepackage[T1]{fontenc}    
\usepackage{hyperref}       
\usepackage{url}            
\usepackage{booktabs}       
\usepackage{amsfonts}       
\usepackage{nicefrac}       
\usepackage{microtype}      
\usepackage{lipsum}
\usepackage{amsmath}
\usepackage{fancyhdr}       
\usepackage{graphicx}       
\graphicspath{{media/}}     
\usepackage[normalem]{ulem}

\title{Attribute Retrieving for Open-Vocabulary Endoscopic Compositional Referring Segmentation}

\author{
  Shun Liu \\
  Virginia Commonwealth University \\
  \texttt{lius24@vcu.edu} \\
  \And
  Nan Xi \\
  Virginia Commonwealth University \\
  \texttt{xin@vcu.edu} \\
  \And
  Yang Liu \\
  King's College London \\
  \texttt{yang.9.liu@kcl.ac.uk} \\
  \And
  Tianyu Luan \\
  University at Buffalo \\
  \texttt{tianyulu@buffalo.edu} \\
  \And
  Xuan Gong \\
  Harvard Medical School \\
  \texttt{xuan\_gong@hms.edu} \\
  \And
  David Doermann \\
  University at Buffalo \\
  \texttt{doermann@buffalo.edu}
}

\begin{document}
\maketitle

\begin{abstract}
Referring Image Segmentation (RIS) aims to segment image regions specified by natural language, enabling fine-grained and controllable visual understanding. Extending RIS to endoscopic imagery, however, presents unique challenges, including scarce high-quality annotations and complex, domain-specific image–text relationships. Although recent vision–language models demonstrate strong cross-domain alignment, they often fail to capture fine-grained textual cues in endoscopic settings, resulting in suboptimal performance and limited generalization. To address these challenges, we introduce \textbf{\textit{ReferEndoscopy}}, a large-scale benchmark for RIS in the endoscopy field. Building on this dataset, we propose the \textbf{Attribute Retrieval-based Endoscopic-RIS (AR-ERIS)} framework for open-vocabulary endoscopic compositional referring segmentation. AR-ERIS leverages attribute retrieval for open-vocabulary endoscopic compositional referring segmentation and is pretrained on the curated ReferEndoscopy dataset, achieving state-of-the-art performance with strong generalization across both simulated and real-world endoscopic data. The dataset and code will be publicly released upon completion of the review process.
\end{abstract}

\section{Introduction}

\begin{figure}
\centering
\includegraphics[width=0.6\linewidth]{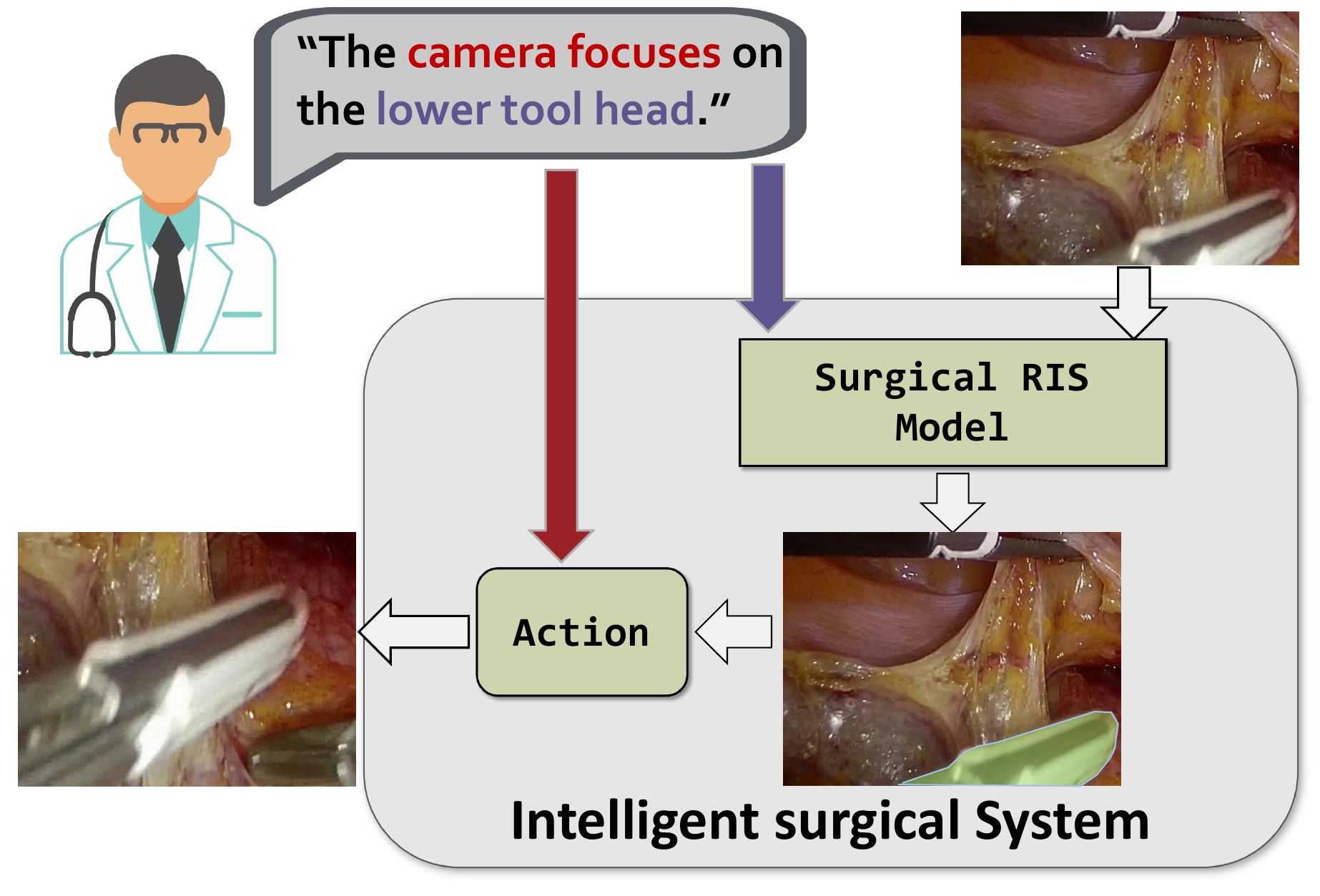}
\caption{\textbf{Application of surgical Referring Image Segmentation~(RIS) in an Intelligent Surgical System.} Upon receiving a user command, the RIS model identifies the specified area of interest, enabling the system to issue precise instructions for further actions within that region.}
\label{fig:intro-application}
\centering
\end{figure}

\begin{figure*}
    \centering
    \includegraphics[width=1\linewidth]{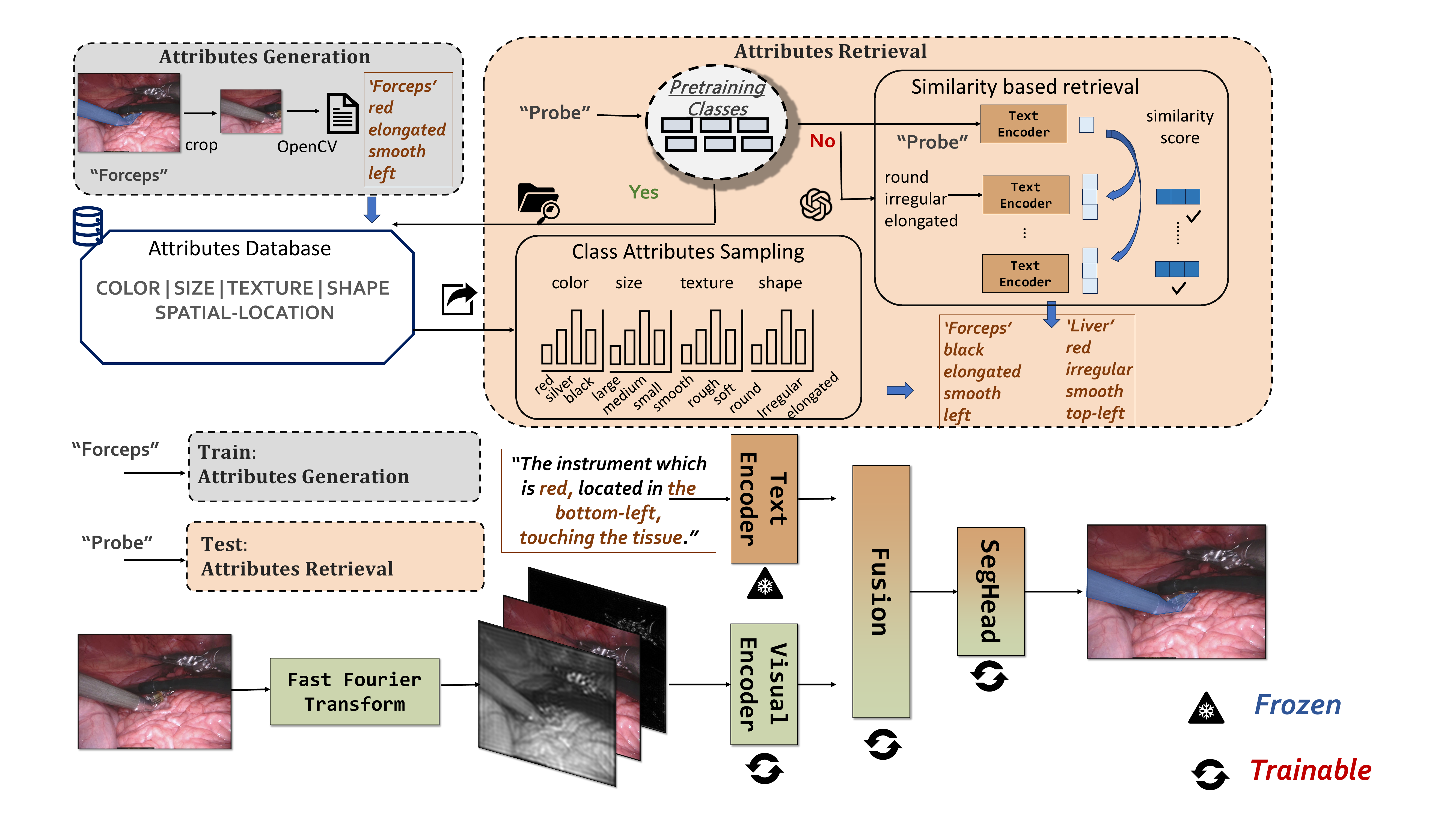}
    \caption{\textbf{The proposed AR-ERIS framework architecture.} Our model comprises a dual model encoder, feature pyramid network (FPN), cross-model attention module, and the projector. Specifically, the text and image modality queries are embedded into separated semantic spaces before entering the feature pyramid network, which generates multi-scale fine-grained information for cross-model alignment. The jointly embedded representations then go through the cross-attention model for further alignment with the residual connections from the original texts; after that, the projectors map the latent vectors into the normal dimensions for pixel classification.}
    \label{fig:enter-label}
\end{figure*}

Analyzing endoscopic images is vital in minimally invasive surgeries and diagnostics by providing real-time, high-resolution visualization of internal structures and organs. Precise identification and segmentation of anatomical regions and surgical instruments are essential to support clinical decision-making and enhance procedural outcomes. While traditional segmentation methods \cite{he2017mask, ronneberger2015u, zhou2019unet++} have achieved notable success in general biomedical image segmentation, endoscopic image segmentation poses unique challenges due to limited detailed annotations, complex textures, occlusions, and dynamic interactions between instruments and tissues.

Referring image segmentation (RIS) \cite{wang2022cris}, which segments object regions based on textual descriptions, has gained recent attention in natural image segmentation. Extending RIS to the medical domain, especially for endoscopic images \cite{wang2022autolaparo, yang2026large, nasirihaghighi2025gynsurg, darjana2025egosurgery}, offers significant potential but remains largely unexplored. As shown in Figure~\ref{fig:intro-application}, in intelligent surgical systems, RIS can be applied to segment specific objects based on textual prompts provided by a user, enhancing the system’s responsiveness to real-time instructions. Unlike static objects in natural images, surgical instruments and tissues in endoscopic videos move and interact frequently, often under challenging conditions such as low lighting, smoke, occlusion, truncation, and blurred boundaries. Efficiently capturing boundary and shape information for open-set categories is thus a critical research question. Current endoscopic image segmentation methods \cite{nasirihaghighi2025gynsurg, darjana2025egosurgery} rely primarily on image-only data within a limited domain, lacking the textual guidance needed for better generalization. This gap highlights the urgent need for a comprehensive text-grounded benchmark and a robust referring segmentation approach. An effective RIS model should accurately segment anatomical structures and instruments in real time based on textual prompts, enhancing the precision and situational awareness of surgical support systems.

Despite recent progress in vision-language segmentation models \cite{lai2024lisa, zou2024segment}, significant challenges remain in compositional referring segmentation for endoscopic images. First, the \textit{fine-grained nature of medical descriptions} requires a model to interpret detailed attributes that go beyond typical object descriptions in natural images, demanding precise recognition of subtle anatomical and instrumental variations. Second, \textit{class imbalance and long-tail distribution} is prevalent in endoscopic datasets, with specific organs or instruments appearing far more frequently than others. Developing a generalized model that performs consistently across varied anatomical contexts and surgical scenarios is crucial.

To address these challenges, we present a new benchmark dataset and a novel framework for endoscopic referring segmentation. The \textbf{\textit{RefEndoscopy}} benchmark comprises 65,964 endoscopic images, 242,055 instruction-grounded masks, and 1,452,330 image–mask –instruction triplets. The benchmark spans a wide range of surgical instruments, anatomical structures, and background variations (e.g., illumination, viewpoints, and endoscope types). The accompanying instructions are carefully designed to encode diverse visual attributes, enabling compositional and fine-grained referring segmentation. We further propose the \textbf{Attribute Retrieval-based Endoscopic-RIS (AR-ERIS)} framework by leveraging attribute retrieval approach to effectively handle the complexities of surgical scene dynamics and the fine-grained specificity required in endoscopic images, enabling open-vocabulary endoscopic compositional referring segmentation. Extensive experiments demonstrate our model’s robustness across diverse surgical scenarios and its capacity to handle inherent class imbalances effectively.

Overall, this work establishes a foundational approach and benchmark for future research in vision-language grounding within the medical domain, with potential applications in enhancing real-time navigation and providing context-aware robotic assistance in minimally invasive procedures. Our contributions are summarized as:
\begin{itemize}
    \item We present ReferEndoscopy, a large-scale, instruction-grounded segmentation benchmark dataset created using an automated visual content recognition pipeline and including samples from diverse endoscopic domains. 
    
    \item We introduce an Attribute Retrieval-based Endoscopic-RIS (AR-ERIS) framework for open-vocabulary compositional referring segmentation. AR-ERIS decomposes RGB inputs into distinct frequency components, capturing diverse attributes including boundary, shape, and texture-specific features. These components are fused with the original visual features through a mixture-of-experts network, enabling adaptive feature selection across varying domains in a learnable, dynamic manner.
    
    \item Extensive experiments demonstrate the effectiveness of our proposed framework, achieving state-of-the-art performance on the ReferEndoscopy benchmark dataset compared to baseline models. Additionally, our model exhibits generalization capabilities in open-vocabulary settings and across previously unseen classes.
\end{itemize}

\section{Task Definition and Benchmark}

\subsection{Task Settings}
Let \( I \in \mathbb{R}^{H \times W \times 3} \) be an input endoscopic image of height \( H \) and width \( W \). \( T \in \mathbb{R}^N \) be a natural language instruction consisting of \( N \) tokens. \( S \in \{0, 1\}^{H \times W} \) be the binary segmentation mask, with \( S_{ij} = 1 \) indicating that pixel \((i, j)\) belongs to the target object as specified by \( T \).
The objective is to learn a function \( f : (I, T) \rightarrow S \) that maps the image-text pair to a segmentation mask \( S \).
For instance, SISVSE~\cite{sisvse} includes 31 fine-grained mask categories for instrument parts visible within endoscopic views. At the same time, Cholecseg8k~\cite{hong2020cholecseg8k} introduces multiple organ and tissue types (e.g., fat, liver, blood, hepatic vein, liver ligament, connective tissue) to expand segmentation in endoscopic imaging. Additionally, datasets like Kvasir-Instrument~\cite{jha2021kvasir} and RoboTool~\cite{robotool} focus on binary segmentation of instruments versus background, further diversifying the scope of \textbf{\textit{ReferEndoscopy}} and integrating these varied categories into a language-image alignment framework during pretraining aims to enable both generalization and robustness in open-vocabulary testing scenarios.

Table \ref{tab:datasets} provides detailed information on the ten datasets that comprise \textbf{\textit{ReferEndoscopy}}. 

\subsection{Attribute-Guided Annotation Pipeline}
\subsubsection{Step\#1: Data Collection, Preprocessing}
We collect and clean data from datasets and benchmarks from various domains to enhance the model generalization ability. In the below, we give detailed profiles of each composition dataset.
\begin{itemize}
\item
\textbf{AutoLaparo}~\cite{wang2022autolaparo}: A laparoscopic dataset focused on multi-instrument segmentation, covering instrument parts within complex surgical environments. It provides annotated masks for multiple instruments and includes challenging cases with occlusions and varying lighting.
\item 
\textbf{DSAD}~\cite{dsad}: The Dresden Surgical Anatomy Dataset includes pixel-level semantic segmentations for eight abdominal organs (e.g., colon, liver, pancreas), the abdominal wall, and two vascular structures, totalling 13,195 laparoscopic images. It supports single-organ and multi-organ segmentation tasks, providing rich annotations that enhance AI model development for complex anatomical structures in laparoscopic surgery.
\item
\textbf{Cholecseg8k}~\cite{hong2020cholecseg8k}: This dataset is designed explicitly for cholecystectomy (gallbladder removal) procedures and contains over 8,000 frames. It includes segmentation masks for organs and tissue types like fat, liver, blood, hepatic veins, liver ligaments, and connective tissues, making it valuable for multi-class organ and tissue segmentation tasks.
\item
\textbf{EndoVis-17/EndoVis-18}~\cite{endo17,endo18}: These are part of the MICCAI EndoVis challenges, focusing on semantic segmentation of instruments in minimally invasive surgery. EndoVis-17 includes multiple instruments with fine-grained part labels, while EndoVis-18 adds complexity with more varied instrument types and occlusions.
\item
\textbf{Kvasir-Instrument}~\cite{jha2021kvasir}: A dataset collected during endoscopic procedures focusing on binary instrument segmentation versus background segmentation. The dataset is suitable for benchmarking binary segmentation models in various endoscopic contexts, specifically gastrointestinal endoscopy.
\item
\textbf{RoboTool}~\cite{robotool}: Designed for robot-assisted surgery, this dataset includes annotated instrument masks in diverse robotic surgery scenes. The masks cover various tools used in procedures, helping to model complex relationships in robot-assisted interventions.
\item 
\textbf{SAR-RARP50}~\cite{sarrarp}: This dataset focuses on robotic-assisted partial nephrectomy, containing video frames from robot-assisted surgeries. It includes detailed annotations for the kidney and surrounding structures, making it useful for organ segmentation and tool-organ interaction analysis.
\item
\textbf{SISVSE}~\cite{sisvse}: The Subtask Surgical Instrument and Visible Structures Endoscopic dataset (SISVSE) provides fine-grained annotations for 31 different instrument parts. This dataset is valuable for instrument-part segmentation and localization within endoscopic scenes, where precise part-level segmentation is essential.
\item
\textbf{LaparoI2I}~\cite{laparoi2i}: This dataset offers diverse surgical scenarios from laparoscopic videos, focusing on instrument segmentation and tool interactions. It provides various instrument masks, allowing generalizable model training across different laparoscopic procedures.
\end{itemize}

\noindent These datasets collectively span a broad spectrum of endoscopic and laparoscopic procedures, providing rich, multi-class annotations that allow the benchmark to support binary and multi-class segmentation tasks in complex, realistic surgical environments.

Some preprocessing steps are taken to refine the image quality. Firstly, we crop the  endoscopy images by filtering out those irrelevant section (surrounding black background) to align with ~\cite{endo17, endo18, jha2021kvasir, hong2020cholecseg8k} and resize them into 448$\times$448 resolution joint pretraining. After that, the collections of image data still lack attributes and instructions. To generate high-quality attributes to explicitly describe the appearances of scene objects in multiple dimensions (\textit{e.g., color/size/texture/location/reference-object, etc.)}, all images are gathered as inputs to the attribute recognition pipeline as shown in the \textbf{Step\#2}.

\subsubsection{Step\#2: Scene Object Attribute Recognition}
To automatically annotate each object with specific attribute values, we employ OpenCV~\cite{bradski2000opencv} image processing tools. The attribute extraction process is structured as follows:

For \textbf{``color"} attribute, We calculate the mean color within the mask using color histograms, yielding values in HSV or RGB space, which are then mapped to basic color labels (e.g., red, silver, black). For \textbf{``size"} attribute, the area \( A \) of the mask is calculated by counting pixels within the mask region:
  \(
  \text{Size} = A = \sum_{(x, y) \in \text{mask}} 1
  \)
which we map to qualitative labels (e.g., small, medium, large); For \textbf{``spatial location"} attribute, the centroid of the mask \( (x_c, y_c) \) provides spatial location:
  \[
  x_c = \frac{\sum_{x} x \cdot I(x, y)}{A}, \quad y_c = \frac{\sum_{y} y \cdot I(x, y)}{A}
  \]
Then, the centroid's relative position within the image frame (e.g., bottom-left corner) is derived accordingly; The \textbf{``reference object/position"} is extracted from masks of the identical images, then calculate the minimal distances between masked region contours, then convert it categorically into text (e.g., touch, near, overlap). Meanwhile, we also extract the relative location (e.g., left/right) between objects to signify their spatial correlations.

\subsubsection{Step\#3: Attribute-Guided Instruction Design}
To achieve robust text-vision alignment for a generalized segmentation model, we design a set of adaptive instructions for each mask by probing the \textit{“attributes”} of visual content. The attributes are defined as the primary characteristics of a region occupied by a mask, including basic information (\textit{color, size, shape, texture, spatial\_location}) and relational information (\textit{reference\_object, relative\_position}). 

Each attribute provides a unique dimension to describe the object semantically. We use a query template engine to generate instructions integrating attribute values into fill-in-the-blank structures. This method allows us to combine attributes into specific prompts that can vary at a detailed level. For example, a prompt like ``\textit{The Needle that is small in size and located in the bottom-left corner}" emphasizes size and spatial information, guiding the model to focus on these aspects. By controlling the number of attributes in each instruction, we construct a coarse-to-fine hierarchy of prompts, where fewer attributes form a general description, and progressively more attributes provide precise localization and distinctive features.

Let \( \mathbf{A} = \{a_1, a_2, \dots, a_n\} \) represent the set of attributes available for a given mask region. For any object \( \mathcal{O} \) in the image, the instruction prompt \( I \) can be generated as:
\[
I_{\mathcal{O}} = \sum_{k=1}^{m} \text{Template}(a_k) \quad \text{where } m \leq n
\]
\( \text{Template}(a_k) \) denotes a template function that incorporates attribute \( a_k \) in the text query, and \( m \) defines the desired level of detail; here, it's the \#attributes.

\subsection{Benchmark Analysis}
Built on top of 10 large-scale datasets, the ``\textbf{\textit{ReferEndoscopy}}" benchmark has 59 categories covering instruments (parts)/organs, the number of images sums up to 65,964, while the total amount of annotated masks is 242,055. To guarantee the fairness and align with the previous work~\cite{hong2020cholecseg8k,endo17,endo18,robotool,wang2022autolaparo,dsad,sisvse,sarrarp,laparoi2i,jha2021kvasir} decently, we split the train/test samples according to their original settings, for the datasets absent of splitting schemes (4 such datasets), 80\% are used for pretraining and the rest preserved for the subsequent testing.

\noindent\textbf{Class Distribution. }Long-tail problem is common in biomedical imaging fields; in our benchmark, we define 'rare classes' as they appear less than 500 samples over 242,055; the ratio is approximately 0.2\%.

\noindent\textbf{Anatomical Class Categorization}
\begin{itemize}
\item
\textbf{Shape-Variable Classes}: These classes, including soft tissues and certain instruments, vary significantly in appearance due to changes in lighting, occlusion, and physical deformation during procedures. Examples are:
   \begin{itemize}
   \item \textbf{Soft Fatty and Connective Tissues}: Fat and connective tissues exhibit flexible and irregular shapes, adapting to pressure or movement, making segmentation challenging. Our model applies adaptive segmentation thresholds, dynamically adjusting its sensitivity to maintain segmentation accuracy.
   \item \textbf{Blood and Other Fluids}: Blood presence is highly variable, especially in surgical areas affected by incisions. 
   \item \textbf{Suction and Suturing Tools}: These tools, such as needles and suction devices, often overlap with other structures and vary in orientation. Our model uses spatial awareness attributes to segment these tools accurately.
   \end{itemize}

\item
\textbf{Fixed Structure Classes}: Fixed anatomical structures and stable instrument parts that exhibit relatively consistent appearances across different surgical scenes. These classes benefit from standard segmentation procedures, providing high precision without frequent adjustments. Examples are:
   \begin{itemize}
   \item \textbf{Organs and Anatomical Structures}: Major liver, spleen, and gallbladder organs have well-defined visual characteristics. 
   \item \textbf{Background and Inner Cavity Walls}: The background and cavity walls provide a spatial context and do not change position or structure. Using stable thresholds, our model segments these areas consistently to establish a clear surgical field.
   \item \textbf{Surgical Instruments}: Key instruments like the bipolar forceps, catheter, and stapler maintain stable appearances, enabling precise segmentation.
   \end{itemize}
\end{itemize}

\noindent\textbf{Instruction Quality.}
Instructions are generated based on the fine-grained attributes in the ``ReferEndoscopy`` benchmark. To formulate logical and descriptive instruction, we leverage the template-based method. Specifically, the attributes of different dimensions are randomly organized. Moreover, we designed four sets of instructions with different \#attributes included. For example, the `naive` instruction has no attribute and becomes the original class name; this aligns with the current methods ~\cite{endo17,endo18,hong2020cholecseg8k,wang2022autolaparo}. Moving beyond this, ``simple``/``medium``/``hard`` indicates one/three/five attributes included, we introduce mainly five attributes in the benchmark to represent different complexities.

\begin{itemize}
\item Naive:
``The [CLS].''
\item Simple: ``The [CLS] that is [$\mathbf{ATTR}_i$].''
\item Medium/Hard:
``The [CLS] with characteristics including [$\mathbf{ATTR}_i$].''
\end{itemize}
where $i\in$ \{`COLOR',`TEXTURE',`LOC',`SHAPE',`SIZE'\}.

\begin{table}[h]
\centering
\caption{\textbf{Dataset Composition Statistics and Training/Testing Splits.} The proposed `\textbf{\textit{ReferEndoscopy}}' benchmark contains ten datasets from both real and virtual domains, with 56 different categories covering instruments/organs/parts. The splitting schemes are subject to the official one.~\cite{endo17,endo18,wang2022autolaparo,hong2020cholecseg8k,dsad, robotool, sarrarp,sisvse,jha2021kvasir, laparoi2i}.}
\label{tab:datasets}
\resizebox{\linewidth}{!}{
    \begin{tabular}{lcccc}
        \toprule
        \textbf{Dataset} & \textbf{Train/Test-set Ratio} & \textbf{\#Train Mask} & \textbf{\#Test Mask} & \textbf{\#Total} \\
        \midrule
        AutoLaparo~\cite{wang2022autolaparo} & 74.89\% & 6254 & 2465 & 8719 \\
        CholecSeg8k~\cite{hong2020cholecseg8k} & 80.00\% & 42952 & 10738 & 53690 \\
        DSAD~\cite{dsad} & 79.88\% & 10551 & 2645 & 13196 \\
        EndoVis-17~\cite{endo17} & 60.00\% & 12250 & 4812 & 17062 \\
        EndoVis-18~\cite{endo18} & 73.33\% & 11369 & 4019 & 15388 \\
        Kvasir-Instrument~\cite{jha2021kvasir} & 80.00\% & 944 & 236 & 1180 \\
        RoboTool~\cite{robotool} & 79.96\% & 811 & 203 & 1014 \\
        SAR-RARP50~\cite{sarrarp} & 80.00\% & 0 & 14228 & 14228 \\
        SISVSE~\cite{sisvse} & 74.83\% & 28873 & 9867 & 38740 \\
        LaparoI2I~\cite{laparoi2i} & 80.00\% & 62251 & 15563 & 77814 \\
        \bottomrule
    \end{tabular}}
\end{table}

\section{Attribute Retrieval-based Endoscopic-RIS (AR-ERIS)}
This section begins with a description of the AR-ERIS framework, followed by an explanation of the associated training and inference processes. Finally, we detail the procedure for generating attributes for the test dataset using the AR-ERIS framework.

\subsection{Frequency-Aware Referring Network}
We introduce the \textbf{F}requency-\textbf{A}ware \textbf{R}eferring Network (FAR-Net) for the AR-ERIS framework to address open-vocabulary endoscopic compositional referring endoscopic segmentation task. The overall framework of AR-ERIS is illustrated in Figure~\ref{fig:enter-label}. FAR-Net builds on the CRIS model~\cite{wang2022cris}, based on CLIPseg~\cite{clipseg} to leverage multimodal contrastive distances for assessing similarity between text and image patches.
In contrast to previous works~\cite{refer,refer1,refer2,refer3}, CRIS employs a cascaded Transformer~\cite{attnis} decoder to transform projected latent vision-language embeddings, generating mask features more effectively.

\noindent\textbf{Frequency-Aware Feature Fusion.}
Most existing methods~\cite{refer,refer1,refer2,refer3,refer4,refer5,refer6,ren2024grounded} encode only the images or image patches for referring segmentation. However, we argue that endoscopic images exhibit significant frequency sensitivity\footnote{Endoscopic images have a simple distribution, with high-frequency regions typically representing the boundaries between instruments and organs and low-frequency regions representing the organ itself.}; thus, jointly perceiving multiple frequency components can significantly enhance image understanding for referring segmentation.
To address this, we propose a feature fusion module called ``Freq-Fusion," which automatically identifies regions with significant pixel value changes (boundaries) and those with minimal changes (textures). 
By separating components based on their frequencies, the model gains more contextual information about class rigidity and deformation rather than treating all components identically. In other words, frequency provides a broad understanding of class appearance, independent of color.

Specifically, given an input image $I \in \mathbb{R}^{B \times 3 \times H \times W}$, where $B$ is the batch size and $H$, $W$ denote the image height and width, we apply a Fourier Transform to decompose the image into high- and low-frequency components. For each channel $I_{i,c}$ of the image, we perform a 2D Fast Fourier Transform (FFT):
\begin{equation}
F_{i,c} = \text{FFT}(I_{i,c}),
\end{equation}

where $F_{i,c}$ represents the frequency domain representation of the image. To separate the high- and low-frequency components, we construct a low-frequency mask $M_{\text{low}}$ and a high-frequency mask $M_{\text{high}}$. Assuming the center of the frequency domain is located at $(c_r, c_c)$, and using radii $r_{\text{low}}$ and $r_{\text{high}}$ to define the low- and high-frequency ranges, these masks are formulated as:
\begin{equation}
M_{\text{low}}(x, y) = \begin{cases} 
      1, & \text{if } \sqrt{(x - c_r)^2 + (y - c_c)^2} < r_{\text{low}} \\
      0, & \text{otherwise}
   \end{cases},
\end{equation}

\begin{equation}
M_{\text{high}} = 1 - M_{\text{low}}.
\end{equation}
Using the frequency masks, we separate the low- and high-frequency components and then apply the inverse Fourier transform to each component to obtain:

\begin{equation}
I_{\text{low}, i, c} = \text{IFFT}(F_{i,c} \cdot M_{\text{low}}),
\end{equation}

\begin{equation}
I_{\text{high}, i, c} = \text{IFFT}(F_{i,c} \cdot M_{\text{high}}),
\end{equation}

where $I_{\text{low}}$ and $I_{\text{high}}$ denote the low- and high-frequency features, respectively. These extracted frequency features are combined with the spatial RGB information to form a composite feature representation, defined as: 
$I_{\text{combined}} = \text{Concat}(I, I_{\text{low}}, I_{\text{high}})$, 
where $I_{\text{combined}} \in \mathbb{R}^{B \times 9 \times H \times W}$. Various fusion strategies, such as naive convolution, bilinear pooling, self-attention, or a mixture of experts, can then be applied to produce the final fused feature map.

Text prompts are embedded using the CLIP~\cite{radfordclip} text encoder and then aligned with the image features $F_{\text{fused}}^{9\times H\times W}$ through cross-modal attention, as described below.

\subsection{Training and Inference}
Multiple loss functions are applied during training to ensure precision and stability in referring segmentation. Specifically, binary cross-entropy (BCE) loss is used to classify pixels in the image accurately. Dice loss penalizes the non-overlapping areas between the predicted masks and the ground truth to distinguish foreground and background contexts further and enhance segmentation accuracy. Additionally, by comparing the high-frequency component $I_{\text{high}}$ with the segmentation mask $M_{\text{mask}}$, the model captures edge details of the target object. The frequency consistency loss is defined as:
\begin{equation}
\mathcal{L}_{\text{freq}} = \| I_{\text{high}} - M_{\text{mask}} \|_2^2.
\end{equation}

\noindent Finally, the total loss is calculated by:
\begin{equation}
\mathcal{L}_{\text{total}} = \mathcal{L}_{\text{BCE}} + \lambda_{\text{freq}} \cdot \mathcal{L}_{\text{freq}} + \lambda_{\text{Dice}} \cdot \mathcal{L}_{\text{Dice}},
\end{equation}
where $\lambda_{\text{Dice}}$, $\lambda_{\text{freq}}$ are empirically set to 0.3, 0.15 by default, respectively.

During inference, FAR-Net takes the class name and image as inputs. To generate accurate instructions based on pre-training knowledge, we first use the $\text{ARM}$ module (which will be introduced in the following subsection) to match the testing class with the pre-training classes.
If the testing class is already included in the model, attributes relevant to the target class—such as color, size, texture, and location—are retrieved from the class-attribute memory bank by selecting the top-k class-attribute pairs.
If the testing class is novel, the $\text{ARM}$ module computes a similarity score between the target class and all pre-training classes and selects the most relevant attributes for instruction generation.
Finally, the instructions are encoded using a standard text encoder and aligned with multi-frequency visual features in the subsequent inference phases.

\subsection{Generation of New Attributes} 
Despite the proposed ReferEndoscopy benchmark's broad coverage, many novel attributes remain uncovered, highlighting a unique challenge. To handle novel attribute segmentation, we design a test-time attribute retrieval module ($\text{ARM}$) which records class-attribute co-occurrences during pre-training.
By dynamically associating attributes with classes throughout large-scale pre-training, $\text{ARM}$ comprehensively represents each class's appearance. Experimental results in Section~\ref{sec-ovt} demonstrate its effectiveness.
At inference, the model retrieves key attributes for specific classes using techniques like top-k sampling to create instructions for novel attributes. There are two cases for attribute generation. First, if an attribute has already been observed, $\text{ARM}$ retrieves it directly from the pre-training memory bank. Second, suppose an attribute is absent in the memory bank. In that case, $\text{ARM}$ uses the CLIP~\cite{radfordclip} encoder to embed the attribute, performs similarity-based matching against the memory bank, and retrieves the most similar class set using top-k sampling, thus incorporating new attributes.

\begin{table*}[t]
\centering
\caption{\textbf{Comparison on \textbf{\textit{ReferEndo}} Task.} The `naive`/`simple`/`medium`/`hard` represent the referring segmentation model or task, where no/one/three/full (five) attributes are used for pretraining/testing in the proposed benchmark. We calculate each subset's mIoU and DICE scores and the whole benchmark. As can be seen from the figure, the model pre-trained with more fine-grained attributes has performance gains over all subsets of \textbf{\textit{ReferEndoscopy}} benchmark. The top-3 scores are in \textbf{bold}, \underline{underlined} and \uuline{double-underlined} respectively.}
\label{tab:model_comparison}
\resizebox{0.98\textwidth}{!}{
    \begin{tabular}{lcc|cc|cc|cc|cc}
        \toprule
        & \multicolumn{2}{c|}{\textbf{Naive}} & \multicolumn{2}{c|}{\textbf{Simple}} & \multicolumn{2}{c|}{\textbf{Medium}} & \multicolumn{2}{c|}{\textbf{Hard}} & \multicolumn{2}{c}{\textbf{Overall}} \\
        \textbf{Model} & \textbf{mIoU (\%)} & \textbf{DICE (\%)} & \textbf{mIoU (\%)} & \textbf{DICE (\%)} & \textbf{mIoU (\%)} & \textbf{DICE (\%)} & \textbf{mIoU (\%)} & \textbf{DICE (\%)} & \textbf{mIoU (\%)} & \textbf{DICE (\%)} \\
        \midrule
        EVF-SAM \cite{zhang2024evf} & 6.84 & 12.80 & 11.03 & 19.87 & 10.61 & 19.18 & 8.19 & 15.14 & 10.47 & 18.96 \\
        GroundedSAM \cite{ren2024grounded} & 10.54 & 19.07 & 9.44 & 17.25 & 8.49 & 15.65 & 9.23 & 16.90 & 9.84 & 17.92 \\
        LAVT \cite{yang2022lavt} & 39.21 & 56.23 & 40.23 & 57.38 & 40.76 & 57.91 & 35.58 & 52.49 & 38.75 & 55.86 \\
        \midrule
        Ours (naive) & 73.76 & 84.90 & 64.54 & 78.45 & 53.34 & 69.57 & 50.06 & 66.72 & 58.38 & 73.72 \\  
        Ours (medium) & \uuline{72.35} & \uuline{84.11} & \underline{73.74} & \underline{84.88} & \textbf{74.80} & \textbf{85.58} & \underline{74.85} & \underline{85.61} & \underline{73.98} & \underline{85.04} \\
        Ours (hard) & 72.62 & 84.14 & 72.09 & 83.78 & 74.37 & 85.30 & 74.23 & 85.21 & 73.45 & 84.69 \\
        Ours (medium w/ $\mathcal{L}_{dice}$) & \underline{73.99} & \underline{85.05} & \textbf{74.81} & \textbf{85.59} & \textbf{75.63} & \textbf{86.12} & \textbf{75.70} & \textbf{86.17} & \textbf{74.64} & \textbf{85.48} \\
        \bottomrule
    \end{tabular}
}
\end{table*}

\section{Experiments}
\subsection{Experimental Setting}
\textbf{Pretraining.} We pre-train the model on a diverse collection of over ten datasets encompassing real and simulated endoscopic images. This broad data exposure enables the model to capture various visual contexts, facilitating generalization across various domains. Pretraining is conducted on a mix of natural language prompts, targeting segmentation masks to help the model learn how textual descriptions correspond to specific image regions.

\noindent\textbf{Implementation Details.} The input images are resized to 448$\times$448, and input sentences are fixed with a maximum length of 77 for our experiments. The feature encoder is originated from CLIP~\cite{radfordclip} to encode $T_{1024d}$ and $V_{512d}$. Feature decoder has three layers, eight heads, 2048 $\text{ffn}$ dimensions, and a 0.1 dropout rate. We set the dice loss weight $W_{dice}$ as 0.3 empirically.
The Adam~\cite{kingma2014adam} optimizer is applied with an initial learning rate $10^{-3}$, then decreased by a factor of $10^{-1}$ at 5-th epoch. The batch size and training epochs are 64 and 20 under 4 NVIDIA 48G A6000 GPUS. 

\subsection{Comparison with referring segmentation models}
To evaluate the effectiveness of our model, we compare its segmentation performance with state-of-the-art methods on the ReferEndoscopy benchmark under four instruction settings, including `naive', `simple', `medium', and `hard', which progressively increase the number of attributes in the referring expressions. Table~\ref{tab:model_comparison} presents comparisons with three leading open-vocabulary segmentation models from the natural image domain~\cite{ren2024grounded, zhang2024evf, yang2022lavt}. Our model consistently outperforms all baselines across all settings in terms of both mean Intersection over Union (mIoU) and Dice scores. In particular, the naive setting achieves the highest performance, with an mIoU of 74.11$\%$, followed by 73.99$\%$ and 72.35$\%$ under more complex settings.

As shown in Table~\ref{tab:model_comparison}, performance varies with the number of attributes in the instruction, which increases the difficulty of vision-language alignment. Specifically, the naive model exhibits a substantial performance drop from 74.11$\%$ to 60.96$\%$ as instruction complexity increases, reflecting its lack of attribute-guided pretraining and limited alignment capability. In contrast, the medium and hard variants demonstrate improved robustness to increasing instruction complexity, achieving mIoU gains of 2.50$\%$ and 1.61$\%$, respectively, under more challenging settings.

\subsection{Comparison with typical segmentation models}
Table~\ref{tab:dataset_model_comparison} compares our method with existing endoscopy segmentation models~\cite{he2022masked, wang2021solo, he2017mask, cheng2022masked, jin2019incorporating, gonzalez2020isinet, isensee2020or}. Notably, these prior approaches are dataset-specific, with models trained and evaluated independently on each narrow task, whereas our approach uses a single unified model evaluated across all datasets. Despite this more challenging setting, our model achieves competitive or superior segmentation performance, demonstrating strong generalization ability across diverse endoscopic scenarios. 
\begin{itemize}
    \item \textbf{Organ vs. Instrument Segmentation:} The results reveal a dichotomy in segmentation challenges. Organs (evaluated in CholecSeg8K) are largely characterized by texture, color, and amorphous boundaries, whereas surgical instruments (evaluated in EndoVis2018 and AutoLaparo) are defined by rigid, high-frequency edges. 
    \item \textbf{Dominance in Soft Tissue (CholecSeg8K):} FAR-Net’s integration of $\mathcal{L}_{dice}$	proves highly effective for soft tissue, effectively mitigating class imbalance. By achieving 92.73$\%$ for liver ligaments and 91.58$\%$ for the abdominal wall, FAR-Net outperforms ISI-Net. This suggests that the text-guided global context allows FAR-Net to distinguish between visually similar tissues (e.g., fat vs. abdominal wall) better than purely visual models. 
    \item \textbf{Edge Preservation in Instruments (EndoVis2018 $\&$ AutoLaparo):} The specialized models initially showed strong baseline performance on surgical tools. However, upon integrating the High-Frequency Loss ($\mathcal{L}_{freq}$), FAR-Net achieves massive gains. The +21.46$\%$ IoU jump for electric hooks and +16.85$\%$ for suction instruments highlights a previous vulnerability in multimodal models: they often produce ``blobby'', low-resolution masks. $\mathcal{L}_{freq}$ forces the decoder to respect the sharp, elongated gradients of metallic tools, allowing FAR-Net to exceed the precision of strictly localized models like OR-UNet.
\end{itemize}

\begin{table*}[t]
\centering
\caption{\textbf{Comparison with Non-Referring segmentation methods.} In this setting, the class names are given. The metrics are ISI-IoU and mcIoU to align with \cite{endo17, endo18}. For other datasets such as AutoLaparo~\cite{wang2022autolaparo} and Cholecseg8k~\cite{hong2020cholecseg8k}. The top-3 scores are in \textbf{bold}, \underline{underlined} and \uuline{double-underlined}.}
\label{tab:dataset_model_comparison}
\resizebox{0.98\textwidth}{!}{
    \begin{tabular}{lcc|cc|cc|cc}
        \toprule
        \textbf{Model} & \multicolumn{2}{c|}{\textbf{EndoVis17}} & \multicolumn{2}{c|}{\textbf{EndoVis18}} & \multicolumn{2}{c|}{\textbf{CholecSeg8K}} & \multicolumn{2}{c}{\textbf{AutoLaparo}} \\
        & \textbf{$IoU_{I}$ (\%)} & \textbf{$IoU_{mc}$ (\%)} & \textbf{$IoU_{I}$ (\%)} & \textbf{$IoU_{mc}$ (\%)} & \textbf{mIoU (\%)} & \textbf{Mean DICE (\%)} & \textbf{mIoU (\%)} & \textbf{Mean DICE (\%)} \\
        \midrule
        MAE~\cite{he2022masked} & - & 44.87 & - & - & - & - & 76.8 & - \\
        SOLO \cite{wang2021solo} & 33.72 & 15.79 & 64.88 & 31.45 & - & - & - & - \\
        MaskRCNN \cite{he2017mask} & 41.77 & 19.59 & \underline{67.94} & 33.66 & - & - & - & - \\
        Mask2Former \cite{cheng2022masked} & 39.84 & 17.78 & 64.69 & 27.67 & 80.7 & 86.9 & 63.2 & 76.4 \\
        MF-TAPNET \cite{jin2019incorporating} & 13.49 & 10.77 & 39.14 & 24.68 & - & - & - & - \\
        ISINet \cite{gonzalez2020isinet} & \textbf{52.20} & 36.79 & \textbf{70.97} & 40.21 & - & - & - & - \\
        OR-UNet~\cite{isensee2020or} & - & - & - & - & 75.4 & 82.7 & 75.3 & 82.1 \\
        \midrule
        Ours (naive) & 46.72 & 53.35 & 60.98 & 78.74 & 83.26 & 90.87 & 75.90 & 86.30 \\
        Ours (medium) & 47.02 & \underline{54.52} & 63.17 & \textbf{80.01} & \uuline{83.64} & \uuline{91.09} & 77.07 & 87.05 \\
        Ours (hard) & 46.09 & \uuline{54.90} & 56.06 & \uuline{79.29} & 83.33 & \underline{90.91} & \textbf{79.10} & \textbf{88.33} \\
        Ours (medium w/ $\mathcal{L}_{dice}$) & \underline{49.06} & \textbf{57.24} & 66.42 & 79.01 & \underline{83.99} & 91.30 & \underline{78.34} & \underline{87.85} \\
        Ours (medium w/ $F^3(\text{MoE})$) & 44.26 & 52.23 & 60.14 & 77.75 & 82.89 & 90.64 & 76.09 & 86.42 \\
        Ours (medium w/ $F^3(\text{MoE})$+$\mathcal{L}_{high-freq}$) & 46.45 & 53.12 & \uuline{67.19} & 78.66 & 83.54 & 91.03 & \uuline{77.19} & \uuline{87.13} \\
        Ours (medium w/ $F^3(\text{MoE})$+$\mathcal{L}_{high-freq}$+$\mathcal{L}_{dice}$) & \uuline{47.99} & 54.17 & 62.62 & \underline{79.35} & \textbf{84.01} & \textbf{91.31} & 77.12 & 87.08 \\
        \bottomrule
    \end{tabular}
}
\end{table*}

\subsection{Generalization on unseen domain}\label{sec-ov}
To comprehensively evaluate generalization, we conduct experiments on an external dataset SAR-RARP50 \cite{sarrarp} that is entirely disjoint from the training data. Table 4 serves as the ultimate stress test for clinical viability. SAR-RARP50 consists of robotic-assisted radical prostatectomy footage, introducing severe domain shifts: different lighting, distinct camera hardware, and novel surgical anatomy compared to the cholecystectomy (gallbladder) datasets likely dominating the pre-training mix. 
\begin{itemize}
    \item \textbf{Outperforming Foundation Models:} Achieving a 9.03$\%$ mIoU lead over GroundedSAM in a zero-shot capacity is substantial. GroundedSAM relies on a pipeline of object detection followed by segmentation, which often fails if the initial text-to-bounding-box step cannot recognize unseen surgical environments. FAR-Net’s end-to-end alignment allows for a more fluid pixel-level understanding, bypassing the bottleneck of strict object detection.
    \item \textbf{Precision Under Ambiguity:} The `hard' model variant achieves a Prec@50 of 20.32$\%$ alongside a 21.32$\%$ mIoU. Prec@50 measures the precision of the mask boundary. In zero-shot scenarios, models typically guess general regions but fail at boundaries. These results indicate that even when FAR-Net encounters an entirely unseen domain, if provided with a highly descriptive ('hard') instruction, it can reliably anchor its segmentation boundaries using the geometric and textural priors established during pre-training. This demonstrates true open-vocabulary generalizability rather than mere dataset memorization.
\end{itemize}

\subsection{Robustness to varying levels of linguistic complexity}
Table 2 evaluates the robustness of referring segmentation models when faced with varying levels of linguistic complexity. The core challenge here is text-visual alignment: as instructions transition from `naive' (e.g., ``the tool'') to `hard' (e.g., ``the metallic grasping forceps interacting with the gallbladder on the left''), traditional models struggle to parse the dominant subject.
\begin{itemize}
    \item \textbf{Degradation of Natural Image Baselines:} We observe a distinct performance drop in state-of-the-art models like GroundedSAM, EVF-SAM, and LAVT as attribute count increases. Specifically, these baselines suffer from "attention dilution," where the model overly attends to modifying adjectives rather than the core surgical target.
    \item \textbf{FAR-Net’s Robustness to Complexity:} In contrast, FAR-Net demonstrates a unique resilience. While our 'naive' pre-trained variant drops from an optimal 74.11$\%$ mIoU to 60.96$\%$ under 'hard' instructions (a natural consequence of lacking attribute-guided priors), our 'medium' and 'hard' variants thrive. By explicitly pre-training on multi-attribute prompts, the model successfully disentangles spatial, morphological, and functional modifiers. The performance gains of 2.50$\%$ and 1.61$\%$ for the 'medium' and 'hard' models respectively validate that FAR-Net leverages complex text as a guiding constraint rather than being confused by it.
    \item \textbf{The Role of the Attribute Retrieval Module (ARM):} The data in Table 2 clearly shows that ARM bridges the gap when textual cues are mismatched with visual evidence. When the instruction is overly simple ('naive' subset) but the visual scene is cluttered, ARM dynamically fetches implicit attributes to stabilize the segmentation, peaking at 75.70$\%$ mIoU when paired with the Dice loss ($\mathcal{L}_{dice}$).
\end{itemize}

\subsection{Ablation Study}

To evaluate the impact of each component, we conduct ablation studies, examining the contribution of each training strategy, loss function, and architectural component.

\noindent\textbf{Dice Loss $\mathcal{L}_{dice}$.}
Under the different instruction complexities in the Table \ref{tab:model_comparison}, the mIoU of `medium' model improves for four settings, with the value of 1.64\%, 1.07\%, 0.83\%, 0.85\% respectively. For the dataset-wise evaluation, we find that integrating $\mathcal{L}_{dice}$ can not only benefit organ type segmentation as well. 
As illustrated in the Table \ref{tab:dataset_model_comparison}, our pre-trained model has expressively results on organ segmentation datasets and benchmark, which has surpassed all baselines across the CholecSeg8K and AutoLaparo datasets, where the organs (abdominal-wall, liver, gastrointestinal tract, fat, connective tissue, blood, gallbladder, hepatic vein and liver ligament) compose the majority of the CholecSeg8K, the top-5 class-wise IoU of our model are 92.73\% (liver ligament), 92.65\% (liver ligament), 91.58\% (abdominal wall), 89.32\% (liver), 88.75\% (fat), respectively; For the AutoLaparo, our pre-trained model performs the best over all baselines and has leading segmentation performances the uterus. 

\noindent\textbf{High Frequency Loss $\mathcal{L}_{freq}$.} To validate how $\mathcal{L}_{freq}$ performs in our framework, we first add the loss item directly into the overall loss by a weighting coefficient of 0.3. The results in the Table \ref{tab:dataset_model_comparison} indicates that introduction of $\mathcal{L}_{freq}$ improves the performances in EndoVis2018 and AutoLaparo, especially for the class $\text{suction instrument}$ (8.86\% IoU gain w/ zero attribute, 16.85\% gain w/ three attributes), 5.81\% w/ full attributes), $\text{electric hook}$ (21.46\% IoU gain w/ zero attribute, 9.36\% gain w/ three attributes, 4.96\% gain w/ full attributes). This phenomenon indicates the role of $\mathcal{L}_{freq}$ in capturing boundary information of the object classes, especially for elongated-shaped instruments.

\noindent\textbf{$\text{ARM}$ Module.} 
To evaluate the effectiveness of this module, we compare the model's performances by gradually increasing the instruction complexity. The results in Table \ref{tab:model_comparison} show the model's performance consistency from pretraining to testing data. Note that testing attributes are generated by $\text{ARM}$ except for `Naive', where no attribute is referred. Experimental results prove that $\text{ARM}$ module can retrieve useful text attributes to aid the model for precise segmentation, when there is no sufficient contextual information. Specifically, the `medium' model has a performance increase from 72.35\% (`naive' subset) to 74.85\% (`hard' subset). This effect is even more prominent when we integrate Dice loss $\mathcal{L}_{dice}$ into the pretraining, the $\text{mIoU}$ has been improved from 73.99\% to 75.70\%, with 1.71\% gains.

\begin{table}[h]
    \centering
    \caption{\textbf{Generalization performance on unseen endoscopic classes from SAR-RARP50 dataset~\cite{sarrarp}.} SAR-RARP50 is used exclusively for evaluation (14,228 image-mask pairs and 85,368 image-mask-instruction pairs). We note there is no overlap between these classes and the training classes.}\label{sec-ovt}
    \label{tab:generalization}
    \begin{tabular}{lcc}
        \toprule
        \textbf{Model} & \textbf{$\text{Prec}@50$} & \textbf{mIoU} \\
        \midrule
        GroundedSAM \cite{ren2024grounded} & 4.85 & 9.25 \\
        \hline
        Ours (naive) & 13.88 & 16.30 \\
        Ours (medium) & 19.46 & \textbf{21.76} \\
        Ours (hard) & \textbf{20.32} & 21.32 \\
        \bottomrule
    \end{tabular}
\end{table}

\section{Related Work}
\noindent \textbf{Endoscopy scene segmentation.} Endoscopy scene segmentation has been an essential task in surgical applications. Previous vision-only works mainly focus on accurately segmenting the whole input video without considering what part the user needs. These works would require the users to pick the part they need from the segmentation results. More recent works use segmentation class names as text inputs so the model can segment the video parts corresponding to the text hint. However, the text inputs are strongly limited to class names, thus creating an evident barrier in real-world applications. Besides, most previous works only used a limited amount of data for training, and their generalizability was not satisfied. In contrast, our method can understand any input text and use the text prompt to locate the user's required segments. Moreover, our model is designed to easily bind several datasets together for training, thus achieving better performance and generalizability. 

\noindent \textbf{RIS in Natural Domain.} Referring image segmentation (RIS) aims to group pixels based on a given natural language description, requiring visual and linguistic understanding. With the introduction of large-scale natural domain datasets such as RefCOCO~\cite{yu2016modeling}, RefCOCO+~\cite{yu2016modeling}, and RefCOCOg~\cite{nagaraja2016modeling}, significant progress has been made in developing RIS methods. Early works~\cite{hu2016segmentation, li2018referring,liu2017recurrent,shi2018key, yang2019cross} primarily employed CNNs to extract visual features and LSTMs for linguistic features. These features were then concatenated and fed into a fully convolutional network (FCN) for segmentation. With the advancement of attention mechanisms and Transformer-based architectures, more powerful cross-modal representation methods~\cite{kim2022restr, lin2021structured, tang2023contrastive, wang2022cris, yang2019dynamic, yang2020graph, yang2020relationship, yang2021bottom, yang2022lavt, chen2022position} have emerged. For example, VLT~\cite{ding2022vlt} enriched global context information through a query generation module, enhancing robustness in feature alignment. LAVT~\cite{yang2022lavt} aligns visual and linguistic representations within the visual backbone using a pixel-word attention module. Recent work CRIS~\cite{wang2022cris} leveraged the strong image-text alignment capability of CLIP~\cite{radford2021learning} to focus on sentence-pixel alignment, while PCAN~\cite{chen2022position} introduced position-aware contrastive alignment to better integrate object location information into visual-linguistic interaction. While RIS in natural domains has benefited from large-scale datasets, the surgical domain lacks similarly extensive datasets, which has presented challenges and limited progress. To address this, we introduce a new large-scale dataset designed explicitly for the surgical domain, providing a solid foundation for future research. We propose a baseline method focusing on efficient feature fusion and alignment using a parameter-efficient approach that leverages pre-trained vision-language models.

\section{Conclusion}
This work introduces ReferEndoscopy, a large-scale benchmark that enables fine-grained and compositional referring segmentation in endoscopic imagery. Building on this benchmark, we propose AR-ERIS, an attribute retrieval-based framework designed for open-vocabulary compositional segmentation. By explicitly modeling attribute-level cues, AR-ERIS improves fine-grained vision–language alignment and demonstrates strong robustness to diverse and complex referring expressions. Extensive experiments show that AR-ERIS achieves state-of-the-art performance and generalizes effectively across both simulated and real-world endoscopic datasets. Looking forward, we will extend RIS to broader multimodal and clinical settings, including integration with surgical reports and robotic systems, may enable richer context-aware understanding and decision support. 

\bibliographystyle{unsrt}
\bibliography{citation}

@inproceedings{hu2016segmentation,
  title={Segmentation from natural language expressions},
  author={Hu, Ronghang and Rohrbach, Marcus and Darrell, Trevor},
  booktitle={Computer Vision--ECCV 2016: 14th European Conference, Amsterdam, The Netherlands, October 11--14, 2016, Proceedings, Part I 14},
  pages={108--124},
  year={2016},
  organization={Springer}
}

@inproceedings{liu2017recurrent,
  title={Recurrent multimodal interaction for referring image segmentation},
  author={Liu, Chenxi and Lin, Zhe and Shen, Xiaohui and Yang, Jimei and Lu, Xin and Yuille, Alan},
  booktitle={Proceedings of the IEEE international conference on computer vision},
  pages={1271--1280},
  year={2017}
}

@inproceedings{li2018referring,
  title={Referring image segmentation via recurrent refinement networks},
  author={Li, Ruiyu and Li, Kaican and Kuo, Yi-Chun and Shu, Michelle and Qi, Xiaojuan and Shen, Xiaoyong and Jia, Jiaya},
  booktitle={Proceedings of the IEEE Conference on Computer Vision and Pattern Recognition},
  pages={5745--5753},
  year={2018}
}

@inproceedings{shi2018key,
  title={Key-word-aware network for referring expression image segmentation},
  author={Shi, Hengcan and Li, Hongliang and Meng, Fanman and Wu, Qingbo},
  booktitle={Proceedings of the European Conference on Computer Vision (ECCV)},
  pages={38--54},
  year={2018}
}

@inproceedings{yang2019cross,
  title={Cross-modal relationship inference for grounding referring expressions},
  author={Yang, Sibei and Li, Guanbin and Yu, Yizhou},
  booktitle={Proceedings of the IEEE/CVF conference on computer vision and pattern recognition},
  pages={4145--4154},
  year={2019}
}

@inproceedings{kim2022restr,
  title={Restr: Convolution-free referring image segmentation using transformers},
  author={Kim, Namyup and Kim, Dongwon and Lan, Cuiling and Zeng, Wenjun and Kwak, Suha},
  booktitle={Proceedings of the IEEE/CVF Conference on Computer Vision and Pattern Recognition},
  pages={18145--18154},
  year={2022}
}

@article{lin2021structured,
  title={Structured attention network for referring image segmentation},
  author={Lin, Liang and Yan, Pengxiang and Xu, Xiaoqian and Yang, Sibei and Zeng, Kun and Li, Guanbin},
  journal={IEEE Transactions on Multimedia},
  volume={24},
  pages={1922--1932},
  year={2021},
  publisher={IEEE}
}

@inproceedings{tang2023contrastive,
  title={Contrastive grouping with transformer for referring image segmentation},
  author={Tang, Jiajin and Zheng, Ge and Shi, Cheng and Yang, Sibei},
  booktitle={Proceedings of the IEEE/CVF Conference on Computer Vision and Pattern Recognition},
  pages={23570--23580},
  year={2023}
}

@inproceedings{wang2022cris,
  title={Cris: Clip-driven referring image segmentation},
  author={Wang, Zhaoqing and Lu, Yu and Li, Qiang and Tao, Xunqiang and Guo, Yandong and Gong, Mingming and Liu, Tongliang},
  booktitle={Proceedings of the IEEE/CVF conference on computer vision and pattern recognition},
  pages={11686--11695},
  year={2022}
}

@inproceedings{yang2019dynamic,
  title={Dynamic graph attention for referring expression comprehension},
  author={Yang, Sibei and Li, Guanbin and Yu, Yizhou},
  booktitle={Proceedings of the IEEE/CVF International Conference on Computer Vision},
  pages={4644--4653},
  year={2019}
}

@inproceedings{yang2020graph,
  title={Graph-structured referring expression reasoning in the wild},
  author={Yang, Sibei and Li, Guanbin and Yu, Yizhou},
  booktitle={Proceedings of the IEEE/CVF conference on computer vision and pattern recognition},
  pages={9952--9961},
  year={2020}
}

@article{yang2020relationship,
  title={Relationship-embedded representation learning for grounding referring expressions},
  author={Yang, Sibei and Li, Guanbin and Yu, Yizhou},
  journal={IEEE Transactions on Pattern Analysis and Machine Intelligence},
  volume={43},
  number={8},
  pages={2765--2779},
  year={2020},
  publisher={IEEE}
}

@inproceedings{yang2021bottom,
  title={Bottom-up shift and reasoning for referring image segmentation},
  author={Yang, Sibei and Xia, Meng and Li, Guanbin and Zhou, Hong-Yu and Yu, Yizhou},
  booktitle={Proceedings of the IEEE/CVF Conference on Computer Vision and Pattern Recognition},
  pages={11266--11275},
  year={2021}
}

@inproceedings{yang2022lavt,
  title={Lavt: Language-aware vision transformer for referring image segmentation},
  author={Yang, Zhao and Wang, Jiaqi and Tang, Yansong and Chen, Kai and Zhao, Hengshuang and Torr, Philip HS},
  booktitle={Proceedings of the IEEE/CVF Conference on Computer Vision and Pattern Recognition},
  pages={18155--18165},
  year={2022}
}

@article{ding2022vlt,
  title={VLT: Vision-language transformer and query generation for referring segmentation},
  author={Ding, Henghui and Liu, Chang and Wang, Suchen and Jiang, Xudong},
  journal={IEEE Transactions on Pattern Analysis and Machine Intelligence},
  volume={45},
  number={6},
  pages={7900--7916},
  year={2022},
  publisher={IEEE}
}

@article{chen2022position,
  title={Position-aware contrastive alignment for referring image segmentation},
  author={Chen, Bo and Hu, Zhiwei and Ji, Zhilong and Bai, Jinfeng and Zuo, Wangmeng},
  journal={arXiv preprint arXiv:2212.13419},
  year={2022}
}

@inproceedings{yu2016modeling,
  title={Modeling context in referring expressions},
  author={Yu, Licheng and Poirson, Patrick and Yang, Shan and Berg, Alexander C and Berg, Tamara L},
  booktitle={Computer Vision--ECCV 2016: 14th European Conference, Amsterdam, The Netherlands, October 11-14, 2016, Proceedings, Part II 14},
  pages={69--85},
  year={2016},
  organization={Springer}
}

@inproceedings{nagaraja2016modeling,
  title={Modeling context between objects for referring expression understanding},
  author={Nagaraja, Varun K and Morariu, Vlad I and Davis, Larry S},
  booktitle={Computer Vision--ECCV 2016: 14th European Conference, Amsterdam, The Netherlands, October 11--14, 2016, Proceedings, Part IV 14},
  pages={792--807},
  year={2016},
  organization={Springer}
}

@inproceedings{radford2021learning,
  title={Learning transferable visual models from natural language supervision},
  author={Radford, Alec and Kim, Jong Wook and Hallacy, Chris and Ramesh, Aditya and Goh, Gabriel and Agarwal, Sandhini and Sastry, Girish and Askell, Amanda and Mishkin, Pamela and Clark, Jack and others},
  booktitle={International conference on machine learning},
  pages={8748--8763},
  year={2021},
  organization={PMLR}
}

@inproceedings{wang2022autolaparo,
  title={Autolaparo: A new dataset of integrated multi-tasks for image-guided surgical automation in laparoscopic hysterectomy},
  author={Wang, Ziyi and Lu, Bo and Long, Yonghao and Zhong, Fangxun and Cheung, Tak-Hong and Dou, Qi and Liu, Yunhui},
  booktitle={International Conference on Medical Image Computing and Computer-Assisted Intervention},
  pages={486--496},
  year={2022},
  organization={Springer}
}

@article{hong2020cholecseg8k,
  title={Cholecseg8k: a semantic segmentation dataset for laparoscopic cholecystectomy based on cholec80},
  author={Hong, W-Y and Kao, C-L and Kuo, Y-H and Wang, J-R and Chang, W-L and Shih, C-S},
  journal={arXiv preprint arXiv:2012.12453},
  year={2020}
}

@misc{endo17,
      title={2017 Robotic Instrument Segmentation Challenge}, 
      author={Max Allan and Alex Shvets and Thomas Kurmann and Zichen Zhang and Rahul Duggal and Yun-Hsuan Su and Nicola Rieke and Iro Laina and Niveditha Kalavakonda and Sebastian Bodenstedt and Luis Herrera and Wenqi Li and Vladimir Iglovikov and Huoling Luo and Jian Yang and Danail Stoyanov and Lena Maier-Hein and Stefanie Speidel and Mahdi Azizian},
      year={2019},
      eprint={1902.06426},
      archivePrefix={arXiv},
      primaryClass={cs.CV},
      url={https://arxiv.org/abs/1902.06426}, 
}

@misc{endo18,
      title={2018 Robotic Scene Segmentation Challenge}, 
      author={Max Allan and Satoshi Kondo and Sebastian Bodenstedt and Stefan Leger and Rahim Kadkhodamohammadi and Imanol Luengo and Felix Fuentes and Evangello Flouty and Ahmed Mohammed and Marius Pedersen and Avinash Kori and Varghese Alex and Ganapathy Krishnamurthi and David Rauber and Robert Mendel and Christoph Palm and Sophia Bano and Guinther Saibro and Chi-Sheng Shih and Hsun-An Chiang and Juntang Zhuang and Junlin Yang and Vladimir Iglovikov and Anton Dobrenkii and Madhu Reddiboina and Anubhav Reddy and Xingtong Liu and Cong Gao and Mathias Unberath and Myeonghyeon Kim and Chanho Kim and Chaewon Kim and Hyejin Kim and Gyeongmin Lee and Ihsan Ullah and Miguel Luna and Sang Hyun Park and Mahdi Azizian and Danail Stoyanov and Lena Maier-Hein and Stefanie Speidel},
      year={2020},
      eprint={2001.11190},
      archivePrefix={arXiv},
      primaryClass={cs.CV},
      url={https://arxiv.org/abs/2001.11190}, 
}

@inproceedings{jha2021kvasir,
  title={Kvasir-instrument: Diagnostic and therapeutic tool segmentation dataset in gastrointestinal endoscopy},
  author={Jha, Debesh and Ali, Sharib and Emanuelsen, Krister and Hicks, Steven A and Thambawita, Vajira and Garcia-Ceja, Enrique and Riegler, Michael A and de Lange, Thomas and Schmidt, Peter T and Johansen, H{\aa}vard D and others},
  booktitle={MultiMedia Modeling: 27th International Conference, MMM 2021, Prague, Czech Republic, June 22--24, 2021, Proceedings, Part II 27},
  pages={218--229},
  year={2021},
  organization={Springer}
}

@article{sarrarp,
  title={Sar-rarp50: Segmentation of surgical instrumentation and action recognition on robot-assisted radical prostatectomy challenge},
  author={Psychogyios, Dimitrios and Colleoni, Emanuele and Van Amsterdam, Beatrice and Li, Chih-Yang and Huang, Shu-Yu and Li, Yuchong and Jia, Fucang and Zou, Baosheng and Wang, Guotai and Liu, Yang and others},
  journal={arXiv preprint arXiv:2401.00496},
  year={2023}
}

@article{robotool,
  author={Garcia-Peraza-Herrera, Luis C. and Fidon, Lucas and D’Ettorre, Claudia and Stoyanov, Danail and Vercauteren, Tom and Ourselin, Sébastien},
  journal={IEEE Transactions on Medical Imaging}, 
  title={Image Compositing for Segmentation of Surgical Tools Without Manual Annotations}, 
  year={2021},
  volume={40},
  number={5},
  pages={1450-1460},
  doi={10.1109/TMI.2021.3057884}
}

@inproceedings{laparoi2i,
  title={Generating large labeled data sets for laparoscopic image processing tasks using unpaired image-to-image translation},
  author={Pfeiffer, Micha and Funke, Isabel and Robu, Maria R and Bodenstedt, Sebastian and Strenger, Leon and Engelhardt, Sandy and Ro{\ss}, Tobias and Clarkson, Matthew J and Gurusamy, Kurinchi and Davidson, Brian R and others},
  booktitle={Medical Image Computing and Computer Assisted Intervention--MICCAI 2019: 22nd International Conference, Shenzhen, China, October 13--17, 2019, Proceedings, Part V 22},
  pages={119--127},
  year={2019},
  organization={Springer}
}

@article{dsad,
  title={The dresden surgical anatomy dataset for abdominal organ segmentation in surgical data science},
  author={Carstens, Matthias and Rinner, Franziska M and Bodenstedt, Sebastian and Jenke, Alexander C and Weitz, J{\"u}rgen and Distler, Marius and Speidel, Stefanie and Kolbinger, Fiona R},
  journal={Scientific Data},
  volume={10},
  number={1},
  pages={1--8},
  year={2023},
  publisher={Nature Publishing Group}
}

@inproceedings{sisvse,
  title={Surgical scene segmentation using semantic image synthesis with a virtual surgery environment},
  author={Yoon, Jihun and Hong, SeulGi and Hong, Seungbum and Lee, Jiwon and Shin, Soyeon and Park, Bokyung and Sung, Nakjun and Yu, Hayeong and Kim, Sungjae and Park, SungHyun and others},
  booktitle={International Conference on Medical Image Computing and Computer-Assisted Intervention},
  pages={551--561},
  year={2022},
  organization={Springer}
}

@inproceedings{radfordclip,
  title={Learning transferable visual models from natural language supervision},
  author={Radford, Alec and Kim, Jong Wook and Hallacy, Chris and Ramesh, Aditya and Goh, Gabriel and Agarwal, Sandhini and Sastry, Girish and Askell, Amanda and Mishkin, Pamela and Clark, Jack and others},
  booktitle={International conference on machine learning},
  pages={8748--8763},
  year={2021},
  organization={PMLR}
}

@article{kingma2014adam,
  title={Adam: A method for stochastic optimization},
  author={Kingma, Diederik P},
  journal={arXiv preprint arXiv:1412.6980},
  year={2014}
}

@article{bradski2000opencv,
  title={OpenCV},
  author={Bradski, Gary and Kaehler, Adrian and others},
  journal={Dr. Dobb’s journal of software tools},
  volume={3},
  number={2},
  year={2000}
}

@inproceedings{clipseg,
  title={Image segmentation using text and image prompts},
  author={L{\"u}ddecke, Timo and Ecker, Alexander},
  booktitle={Proceedings of the IEEE/CVF conference on computer vision and pattern recognition},
  pages={7086--7096},
  year={2022}
}

@article{attnis,
  title={Attention is all you need},
  author={Vaswani, A},
  journal={Advances in Neural Information Processing Systems},
  year={2017}
}

@inproceedings{gonzalez2020isinet,
  title={Isinet: an instance-based approach for surgical instrument segmentation},
  author={Gonz{\'a}lez, Cristina and Bravo-S{\'a}nchez, Laura and Arbelaez, Pablo},
  booktitle={International Conference on Medical Image Computing and Computer-Assisted Intervention},
  pages={595--605},
  year={2020},
  organization={Springer}
}

@inproceedings{jin2019incorporating,
  title={Incorporating temporal prior from motion flow for instrument segmentation in minimally invasive surgery video},
  author={Jin, Yueming and Cheng, Keyun and Dou, Qi and Heng, Pheng-Ann},
  booktitle={Medical Image Computing and Computer Assisted Intervention--MICCAI 2019: 22nd International Conference, Shenzhen, China, October 13--17, 2019, Proceedings, Part V 22},
  pages={440--448},
  year={2019},
  organization={Springer}
}

@article{zhou2019unet++,
  title={Unet++: Redesigning skip connections to exploit multiscale features in image segmentation},
  author={Zhou, Zongwei and Siddiquee, Md Mahfuzur Rahman and Tajbakhsh, Nima and Liang, Jianming},
  journal={IEEE transactions on medical imaging},
  volume={39},
  number={6},
  pages={1856--1867},
  year={2019},
  publisher={IEEE}
}

@inproceedings{ronneberger2015u,
  title={U-net: Convolutional networks for biomedical image segmentation},
  author={Ronneberger, Olaf and Fischer, Philipp and Brox, Thomas},
  booktitle={Medical image computing and computer-assisted intervention--MICCAI 2015: 18th international conference, Munich, Germany, October 5-9, 2015, proceedings, part III 18},
  pages={234--241},
  year={2015},
  organization={Springer}
}

@inproceedings{he2022masked,
  title={Masked autoencoders are scalable vision learners},
  author={He, Kaiming and Chen, Xinlei and Xie, Saining and Li, Yanghao and Doll{\'a}r, Piotr and Girshick, Ross},
  booktitle={Proceedings of the IEEE/CVF conference on computer vision and pattern recognition},
  pages={16000--16009},
  year={2022}
}

@inproceedings{he2017mask,
  title={Mask r-cnn},
  author={He, Kaiming and Gkioxari, Georgia and Doll{\'a}r, Piotr and Girshick, Ross},
  booktitle={Proceedings of the IEEE international conference on computer vision},
  pages={2961--2969},
  year={2017}
}

@article{isensee2020or,
  title={Or-unet: an optimized robust residual u-net for instrument segmentation in endoscopic images},
  author={Isensee, Fabian and Maier-Hein, Klaus H},
  journal={arXiv preprint arXiv:2004.12668},
  year={2020}
}

@inproceedings{cheng2022masked,
  title={Masked-attention mask transformer for universal image segmentation},
  author={Cheng, Bowen and Misra, Ishan and Schwing, Alexander G and Kirillov, Alexander and Girdhar, Rohit},
  booktitle={Proceedings of the IEEE/CVF conference on computer vision and pattern recognition},
  pages={1290--1299},
  year={2022}
}

@inproceedings{lai2024lisa,
  title={Lisa: Reasoning segmentation via large language model},
  author={Lai, Xin and Tian, Zhuotao and Chen, Yukang and Li, Yanwei and Yuan, Yuhui and Liu, Shu and Jia, Jiaya},
  booktitle={Proceedings of the IEEE/CVF Conference on Computer Vision and Pattern Recognition},
  pages={9579--9589},
  year={2024}
}

@article{zou2024segment,
  title={Segment everything everywhere all at once},
  author={Zou, Xueyan and Yang, Jianwei and Zhang, Hao and Li, Feng and Li, Linjie and Wang, Jianfeng and Wang, Lijuan and Gao, Jianfeng and Lee, Yong Jae},
  journal={Advances in Neural Information Processing Systems},
  volume={36},
  year={2024}
}

@article{ren2024grounded,
  title={Grounded sam: Assembling open-world models for diverse visual tasks},
  author={Ren, Tianhe and Liu, Shilong and Zeng, Ailing and Lin, Jing and Li, Kunchang and Cao, He and Chen, Jiayu and Huang, Xinyu and Chen, Yukang and Yan, Feng and others},
  journal={arXiv preprint arXiv:2401.14159},
  year={2024}
}

@article{zhang2024evf,
  title={Evf-sam: Early vision-language fusion for text-prompted segment anything model},
  author={Zhang, Yuxuan and Cheng, Tianheng and Hu, Rui and Liu, Lei and Liu, Heng and Ran, Longjin and Chen, Xiaoxin and Liu, Wenyu and Wang, Xinggang},
  journal={arXiv preprint arXiv:2406.20076},
  year={2024}
}

@article{wang2021solo,
  title={Solo: A simple framework for instance segmentation},
  author={Wang, Xinlong and Zhang, Rufeng and Shen, Chunhua and Kong, Tao and Li, Lei},
  journal={IEEE transactions on pattern analysis and machine intelligence},
  volume={44},
  number={11},
  pages={8587--8601},
  year={2021},
  publisher={IEEE}
}

@inproceedings{refer,
  title={Locate then segment: A strong pipeline for referring image segmentation},
  author={Jing, Ya and Kong, Tao and Wang, Wei and Wang, Liang and Li, Lei and Tan, Tieniu},
  booktitle={Proceedings of the IEEE/CVF Conference on Computer Vision and Pattern Recognition},
  pages={9858--9867},
  year={2021}
}

@inproceedings{refer1,
  title={Vision-language transformer and query generation for referring segmentation},
  author={Ding, Henghui and Liu, Chang and Wang, Suchen and Jiang, Xudong},
  booktitle={Proceedings of the IEEE/CVF International Conference on Computer Vision},
  pages={16321--16330},
  year={2021}
}

@inproceedings{refer2,
  title={Language as queries for referring video object segmentation},
  author={Wu, Jiannan and Jiang, Yi and Sun, Peize and Yuan, Zehuan and Luo, Ping},
  booktitle={Proceedings of the IEEE/CVF Conference on Computer Vision and Pattern Recognition},
  pages={4974--4984},
  year={2022}
}

@article{refer3,
  title={Cross-modal progressive comprehension for referring segmentation},
  author={Liu, Si and Hui, Tianrui and Huang, Shaofei and Wei, Yunchao and Li, Bo and Li, Guanbin},
  journal={IEEE Transactions on Pattern Analysis and Machine Intelligence},
  volume={44},
  number={9},
  pages={4761--4775},
  year={2021},
  publisher={IEEE}
}

@inproceedings{refer4,
  title={Learning to segment every referring object point by point},
  author={Qu, Mengxue and Wu, Yu and Wei, Yunchao and Liu, Wu and Liang, Xiaodan and Zhao, Yao},
  booktitle={Proceedings of the IEEE/CVF Conference on Computer Vision and Pattern Recognition},
  pages={3021--3030},
  year={2023}
}

@inproceedings{refer5,
  title={Restr: Convolution-free referring image segmentation using transformers},
  author={Kim, Namyup and Kim, Dongwon and Lan, Cuiling and Zeng, Wenjun and Kwak, Suha},
  booktitle={Proceedings of the IEEE/CVF Conference on Computer Vision and Pattern Recognition},
  pages={18145--18154},
  year={2022}
}

@inproceedings{refer6,
  title={End-to-end referring video object segmentation with multimodal transformers},
  author={Botach, Adam and Zheltonozhskii, Evgenii and Baskin, Chaim},
  booktitle={Proceedings of the IEEE/CVF Conference on Computer Vision and Pattern Recognition},
  pages={4985--4995},
  year={2022}
}

@article{yang2026large,
  title={Large-scale self-supervised video foundation model for intelligent surgery},
  author={Yang, Shu and Zhou, Fengtao and Mayer, Leon and Huang, Fuxiang and Chen, Yiliang and Wang, Yihui and He, Sunan and Nie, Yuxiang and Wang, Xi and Jin, Yueming and others},
  journal={npj Digital Medicine},
  year={2026},
  publisher={Nature Publishing Group UK London}
}

@inproceedings{nasirihaghighi2025gynsurg,
  title={Gynsurg: A comprehensive gynecology laparoscopic surgery dataset},
  author={Nasirihaghighi, Sahar and Ghamsarian, Negin and Peschek, Leonie and Munari, Matteo and Husslein, Heinrich and Sznitman, Raphael and Schoeffmann, Klaus},
  booktitle={Proceedings of the 33rd ACM International Conference on Multimedia},
  pages={13141--13147},
  year={2025}
}

@article{darjana2025egosurgery,
  title={EgoSurgery-HTS: A Dataset for Egocentric Hand--Tool Segmentation in Open Surgery Videos},
  author={Darjana, Nathan and Fujii, Ryo and Saito, Hideo and Kajita, Hiroki},
  journal={Healthcare Technology Letters},
  volume={12},
  number={1},
  pages={e70049},
  year={2025},
  publisher={Wiley Online Library}
}

\end{document}